\documentclass[sigconf,screen]{acmart}
\usepackage{graphicx,amsfonts,comment,subcaption,amsmath,multirow}
\usepackage[linesnumbered,ruled]{algorithm2e}

\AtBeginDocument{%
  \providecommand\BibTeX{{%
    \normalfont B\kern-0.5em{\scshape i\kern-0.25em b}\kern-0.8em\TeX}}}

\copyrightyear{2021}
\acmYear{2021}
\setcopyright{none}

\acmConference[arXiv]{arXiv}{CoRR}{2021}
\acmPrice{0}
\acmISBN{NA}
\let\vec\mathbf
\begin{document}

\title[Context-aware demand prediction in bike sharing systems]{Context-aware demand prediction in bike sharing systems: incorporating spatial, meteorological and calendrical context} 

\author{Cláudio Sardinha}
\affiliation{\institution{INESC-ID and IST, Univ. de Lisboa}}
\email{claudio.sardinha@tecnico.ulisboa.pt}
\orcid{1234-5678-9012}

\author{Anna C. Finamore}
\affiliation{\institution{INESC-ID and IST, Univ. de Lisboa}}
\email{anna.couto@tecnico.ulisboa.pt}
\orcid{0000-0001-7182-2683}

\author{Rui Henriques}
\affiliation{%
 \institution{INESC-ID and IST, Univ. de Lisboa}}
\email{rmch@tecnico.ulisboa.pt}
\orcid{0000-0002-3993-0171}

\renewcommand{\shortauthors}{Sardinha et al.}

\begin{abstract}
Bike sharing demand is increasing in large cities worldwide. %
The proper functioning of bike-sharing systems is, nevertheless, dependent on a balanced geographical distribution of bicycles throughout a day. 
In this context, understanding the spatiotemporal distribution of check-ins and check-outs is key for station balancing and bike relocation initiatives. Still, recent contributions from deep learning and distance-based predictors show limited success on forecasting bike sharing demand. This consistent observation is hypothesized to be driven by: i) the strong dependence between demand and the meteorological and situational context of stations; and ii) the absence of spatial awareness as most predictors are unable to model the effects of high-low station load on nearby stations. 

This work proposes a comprehensive set of new principles to incorporate both historical and prospective sources of spatial, meteorological, situational and calendrical context in predictive models of station demand. To this end, a new recurrent neural network layering composed by serial long-short term memory (LSTM) components  is proposed with two major contributions: i) the feeding of multivariate time series masks produced from historical context data at the input layer, and ii) the time-dependent regularization of the forecasted time series using prospective context data. 
This work further assesses the impact of incorporating different sources of context, showing the relevance of the proposed principles for the community even though not all improvements from the context-aware predictors yield statistical significance. 
\end{abstract}


\maketitle

\section{Introduction}
\label{intro-sec}

Transportation dynamics are changing in large cities worldwide \cite{creutzig2012decarbonizing, nieuwenhuijsen2020urban, karlsson2020climate}. Modes of shared mobility are rising in popularity, particularly bike sharing modes propelled by structural shifts in the culture and cycling infrastructures of urban systems \cite{shaheen2011worldwide, campbell2017sharing}. Worldwide, the total number of bikes is estimated to have increased from 700,000 bikes in 2014 towards over 18 million in 2018 \cite{statsbss}. To reduce the carbon fingerprint and meet this demand, over 1600 Bike Sharing Systems (BSS) are now operating and rapidly expanding \cite{statsbss} as they offer a reliable, low-cost and environmental-friendly mode of short-distance transportation. 
Most public BSS support cycling traffic along a network of docking stations, as opposed to dockless bike sharing systems \cite{shaheen2020sharing}.

In contrast with other modes of transportation, the operation and planning of public BSS in metropolitan areas are hindered by unique challenges. First, the demand is unevenly distributed both geographically and temporally, with stations in residence and office areas being either frequently full or empty during peak hours, thus preventing local check-ins and check-outs and consequently hampering user trust 
\cite{frade2015bike, ren2020improved}. Second, bike sharing demand is affected by significant externalities, including daily variations on the users' profile and endeavors impacting their preference towards a given mode of mobility
\cite{kim2018investigation}. Balancing initiatives, including the ongoing bike relocation or dynamic user incentives for taking specific routes along certain time windows can be placed to counter-act this effect \cite{ren2020improved}. Yet, their efficacy is largely dependent on the ability to model and forecast the demand of BSS stations. 

In this context, predicting bike sharing demand at fine spatial and temporal levels is essential for the proper operation and planning of public BSS. Despite the large attention placed by the communities of urban computing, machine learning, and intelligent transportation systems on this task \cite{lin2018predicting, li2019learning, lozano2018multi, chen2020predicting}, BSS demand prediction is recognizably a difficult task. State-of-the-art predictors, whether grounded on classical, distance-based or neural processing principles, show unexpectedly large forecasting errors in comparison with other transportation modes \cite{leitecontext}. In addition to the aforementioned challenges, the difficulty on predicting demand is linked with the fact that bike sharing demand is significantly dependent on:
\begin{itemize}
    \item \vskip -.1cm the \textit{situational and calendrical context}. The occurrence of public events and the presence of calendar-driven context -- including academic breaks, festivities and weekly-monthly-yearly seasonal factors -- affect demand, and can significantly alter expectations on station load \cite{leitecontext};
    \item \vskip .15cm the \textit{meteorological context}. Weather factors, including precipitation, humidity and significant deviations to the perceived temperature are known to condition user decisions \cite{kim2018investigation};
    \item \vskip .15cm \textit{spatial context} (station-wise interdependencies). States of full and empty station load not only hamper the demand analysis on the stations subjected to check-in and check-out impediments, but also affect the demand of nearby stations, which will receive an indirect increase in the demand for bike check-ins and check-outs that should be separated from their true demand under normal conditions.  
    In this context, variables such as distance between stations, station capacity and other interdependence factors impact demand \cite{fishman2014barriers, frade2015bike}.
\end{itemize}
\vskip -0.1cm

Recent contributions have been proposed to consider calendrical constraints \cite{kim2018investigation}; offer meteorological-based corrections by either segmenting data 
\cite{cerqueira} or extending the learning process \cite{ashqar2019modeling}; or model spatial dependencies using influence factors \cite{tomaras2018modeling} or graph neural networks \cite{pan2019urban}. Despite their relevance, 
state-of-the-art contributions for predictive tasks generally fail to model the joint impact that these multiple sources of context exert on bike sharing demand. 

In addition, existing work generally fails to separate 
the important role of both historical and prospective sources of context. For instance, historical weather data are important to assert the true impact of weather variables in demand, while prospective weather forecasts essential to adjust predictions. Context-sensitive models generally tackle one of these two modes, either historical sources of context to consistently remove context-dependent factors from predictions or prospective sources to embed context-dependent factors throughout predictions.

This work proposes a predictive deep learning approach that is able to incorporate heterogeneous sources of spatial, meteorological and calendrical context -- both historical and prospective -- into the time series forecasting task. To this end, we propose a new class of recurrent neural layering for context-sensitive demand prediction. 
Motivated by the solid performance of long-short term memory networks (LSTMs) in traffic data analysis, 
this work proposes a serial composition of recurrent components using two major principles:
\begin{enumerate}
\item \vskip -0.06cm 
to take advantage of the inherent ability of LSTMs to learn from multivariate time series data in order to model correlations between demand variables and an arbitrarily-high number of context variables derived from the available historical context. To this end, a set of simplistic yet effective masking procedures are proposed; 
\item to include an additional LSTM or gated recurrent unit (GRU) layering for the time-dependent regularization of the forecasted signal using prospective context data collected along the horizon of prediction.
\end{enumerate}
\vskip -0.06cm

This proposal is motivated and assessed in the context of BSS demand prediction given its paramount relevance in this domain. In particular, our work further introduces a study case -- GIRA, the public BSS in the Lisbon city (77 stations, approximately 700 bikes and 700 to 1500 daily trips) -- and evaluates the impact of incorporating available sources of context in the ability to predict the demand along the GIRA's bike sharing system. 
The gathered results 
show the role of specific sources of historical and prospective context data in shaping demand forecasts, 
motivating the 
relevance of the proposed principles to support BSS balancing and planning. 

The major contributions of this work are three-fold: i) extension of LSTM-based predictive models to incorporate heterogeneous sources of context data from masking principles; ii) serial composition of LSTM components for the time-dependent regularization of the forecasted signal based on prospective context data; and iii)  comprehensive assessment of the impact of the proposed context-awareness forecasters against state-of-the-art forecasters at different spatial and temporal granularities.

The manuscript is structured as follows. Section \ref{back-sec} provides essential background on bike sharing systems, time series forecasting, and urban sources of context. Section \ref{rw-sec} surveys relevant work on context-sensitive prediction. Section \ref{sol-sec} introduces the proposed approach, describing principles on how to incorporate historical and prospective context data within the learning process. Section \ref{res-sec} 
assesses the impact of these principles. 
Finally, concluding remarks and implications 
are synthesized in section \ref{disc-sec}.
\section{Background}
\label{back-sec}

\subsection*{Bike sharing systems}

Bike sharing systems (BSS) allow users to access bicycles on an as-needed basis using docking or dockless infrastructure. BSS are constituted by two entities: agents (users) and the environment (bikes and stations). A bike station is described by the number of docks (\textit{station capacity}) and number of docked bikes (\textit{station load}). Agents change the environment by performing \textit{check-out} acts (picking-up bikes) and \textit{check-in} acts (dropping-off bikes). 

In the context of this work, \textit{bike demand} is defined as the number of check-in and check-out acts that should be supported at a specific station in a given time to fill the user's needs. 

\textit{BSS traffic data} are the monitored check-ins and check-outs along the network of stations per user, accompanied by information on the station capacity and load. 

\textit{Context data} are data produced from alternative urban sources of interest, including public events (situational), weather records \mbox{(meteorological), seasonal determinants (calendrical), among others.}

Given a bike sharing system, the \textbf{target task} in this work is to \textit{forecast bike demand at desirable spatial and temporal levels from available traffic and context data}.

\subsection*{Time series modeling and forecasting}

Traffic and context data are generally represented as \textit{time series}, a set of ordered observations $\vec{x}_{1..T}=(\vec{x}_{1},...,\vec{x}_{T})$, each observation $\vec{x}_{t}$ being recorded at a specific time point $t$. Time series can be \textit{univariate}, $\vec{x}_{t}\in\mathbb{R}$, or \textit{multivariate}, $\vec{x}_{t}\in\mathbb{R}^m$, where $m>1$ is the multivariate order (number of variables). 

Given a time series $\vec{x}$, the \textit{modeling task} aims at finding an abstraction that describes $\vec{x}$, while the \textbf{forecasting task} aims to estimate $h$ upcoming observations, $\vec{x}_{T+1..T+h}$, from available observations $\vec{x}_{1..T}$. 
\textit{Descriptive} and \textit{predictive} tasks typically aim at respectively modeling and forecasting a single variable (the target variable). 

In multivariate data settings, non-target variables are used to guide the learning of descriptive and predictive models.

Classical approaches for time series analysis generally rely on statistical principles, including decomposition, auto-regression, differencing and exponential smoothing operations. Time series can be decomposed into trend, seasonal, cyclical, and irregular components using additive or multiplicative models \cite{jain2001state}. Although these models are inherently descriptive in nature, the components can be projected along a time horizon for predictive ends. Alternatively, auto-regressive 
and moving-average models can be combined to either describe or forecast stationary time series \cite{wei2006time}. These models can be extended with simplistic differencing operations 
for dealing with non-stationarity, and their learning guided in the presence of seasonal terms.  Classical alternatives use exponential smoothing principles to describe and predict time series \cite{wei2006time}. Holt-Winters offers a form of triple exponential smoothing (at level $\ell_{t}$, trend $b_{t}$ and seasonal $s_{t}$ components). According to an additive model,
\vskip -0.3cm
\footnotesize
\begin{align}
    \hat{\vec{x}}_{t+h} &= (\ell_{t} + \vec{h}b_{t})s_{t+h-d(k+1)} \text{ where} \\
    \ell_{t} &= \alpha \frac{\vec{x}_{t}}{s_{t-d}} + (1 - \alpha)(\ell_{t-1} + b_{t-1}),\\
    b_{t} &= \beta^*(\ell_{t}-\ell_{t-1}) + (1 - \beta^*)b_{t-1},\\
    s_{t} &= \gamma \frac{\vec{x}_{t}}{(\ell_{t-1} + b_{t-1})} + (1 - \gamma)s_{t-d}.
\end{align}


\normalsize 
Advances from machine learning aim at mitigating the challenges faced by classical approaches. In this context, time series are generally segmented to compose a dataset that guides the learning of the target descriptive and predictive models under a specific loss criteria. Among the wide-diversity of contributions on (multivariate) time series analysis, two major groups of approaches are here highlighted. 
First, \textit{distance-based approaches} for time series description and prediction that rely on similarities between (multivariate) time series (\textit{lazy learning}) and expectations (\textit{barycenter computation}) \cite{cai2016spatiotemporal}. 
Second, \textit{neural network approaches} rely on the composition of simple linear functions (neurons) to learn complex mappings. In the context of time series analysis, the mapping can either be descriptive (e.g. \textit{auto-encoders}) or predictive (e.g. \textit{regressors}) \cite{bao2017deep}.  

\textit{Recurrent neural networks} (RNNs) capture the temporal dependencies within a time series, subjecting each observation to a neural network and feeding the result to the successor observation (hence recurrent), 
\vskip -0.6cm
\begin{equation}
    \vec{h}_{t} = f (\vec{h}_{t-1} , \vec{x}_{t}),
\end{equation}

\noindent where $\vec{h}_{t}$ is the output produced at time point $t$ and $f$ is the neural function. As neurons are linear functions of inputs $w_i$, $\sum_i w_i$,  
\vskip -0.3cm

\begin{equation}
    \vec{h}_{t} = \text{tanh} ( \vec{W}_{x,t-1} \vec{h}_{t-1} + \vec{W}_{t} \vec{h}_{x,t} ),
\end{equation}

\noindent where 
$\vec{W}_{t}$ is a matrix of input weights at state $t$ and \textit{tanh} is the hyperbolic tangent (activation) function. The output state is defined as $\vec{W}_{y,t} \vec{h}_{t}$, where $\vec{W}_{y,t}$ is the set weights at the output state. 

RNNs suffer from memory loss, the well-studied vanishing gradient problem \cite{hochreiter1998vanishing}, being generally unable to detect seasonal time series aspects. \textit{Long Short Term Memory networks} (LSTMs) are extended RNNs able to learn long-term dependencies. LSTMs are composed by a memory unit (cell) and three regulators (gates) that control the flow of information inside the LSTM -- the preserved information (forget gate), the updatable inputs (input gate) and the candidate outputs (output gate). \textit{Gated Recurrent Units} (GRUs) exclude the output gate, providing a simpler network that generally yield improvements in denoising tasks \cite{chung2014empirical}.

\subsection*{Context} 
Diverse sources of urban context data are known to have soft-to-strong correlations with traffic dynamics \cite{cerqueira}, suggesting their relevance to guide descriptive and predictive learning tasks. These sources can be divided according to: i) \textit{calendrical} sources, providing relevant information associated with academic and holiday periods, as well as daily, weekly, monthly and yearly seasonality; ii) \textit{situational} sources, including historical and prospective public events (such as conventions, festivals, concerts, sport events) and road interdictions (including construction works); 
iii) \textit{meteorological} sources, comprising observed and forecasted weather variables (such as perceived temperature, humidity, wind intensity, precipitation, visibility); and iv) \textit{spatial} sources capturing geographical dependencies with potential value for traffic data analysis \cite{casas2019spatially}.

\section{Related work}
\label{rw-sec}

Recent attention has been paid on how to incorporate context factors to understand mobility dynamics and support traffic data analysis \cite{cerqueira,leitecontext}. These factors are typically divided on whether they can be planned \cite{latoski2003managing} -- including football matches, concerts, festivals, construction works, urban planning -- or not \cite{Soua2016BigdatageneratedTF} -- weather, air quality, traffic accidents, emergencies. The former factors are often referred as planned special events (PSE) \cite{latoski2003managing}. Some of the challenges of integrating spatiotemporal context and its role in developing smart cities are discussed by Sagl et al. \cite{sagl2015contextual}. 

In this paper, we are interested in three large family of studies: i) studies that assess the correlation impact of context factors; ii) studies that place general principles on how to use context for guiding forecasting tasks; and iii) deep learning studies able to incorporate context factors within neural network learning.

In the first category, Tomaras et al. \cite{tomaras2018modeling} proposed the use of a metric, influence factor, to measure the impact of social events in the demand of nearby bike stations. The influence factor is a ratio between the sum of the drop-off and pick-up when an event happens, and drop-off and pick-up in a typical day. This factor can be used as a correction factor for descriptive and predictive models. Kwoczek et al. \cite{pseprediction} proposed a method to predict and visualize traffic congestions caused by planned special events. 
Public events are generally characterized by two waves of congestion: people arriving and leaving the event. Recognizing the difficulty of estimating the impact of these waves (the popularity of event), the authors developed a system to predict the waves using nearest neighbors from past PSEs and showed that event-sensitive predictions yield improvements. 
Soltani et al. \cite{soltani2019bikesharing} conducted a web-based survey from public BSS and private BSS  users (OfO and O’Bike) at Adelaide, Australia, and explored key findings to support policy makers. 
Context factors, including relatively low car dependency; young user profile; large share of students, visitors, and non-australian residents; safety concerns; and facility conditions were shown to highly impact demand. 
Yang et al. \cite{yang2019spatiotemporal} studied the impact that a new metro line has in dockless bike sharing systems. To this end, they considered around 80,000 dockless bikes in Nanchang, China to capture changes in behaviours and patterns in the urban flow. 

In the second category, studies that have considered context to enrich forecasting, 
Gallop et al. \cite{gallop2012seasonal} explore complex serial correlation patterns between weather and bike traffic and use this effect to adjust the error terms of the classic autoregressive integrated moving average (ARIMA) models. They further suggest that this correction can be used to affect historical data in an effort to create context-independent models that facilitate traffic data analysis tasks. El-Assi et al. \cite{torronto2017effects} provide a multi-level model considering the impact of land use, built environment, and weather measures on bike share ridership. Similarly, Tran et al. \cite{tran2015modeling} consider the problem of predicting bike-sharing system flow. To this end, they propose the use of a regression model and further consider the effects from five categories of context variables: public transport, socio-economic, topographic, bike-sharing network, and leisure variables. Ashqar et al. \cite{ashqar2019modeling} consider a data-fusion approach towards the analysis of context-enriched bike demand. They propose the use of random forests to rank context predictors and consider them to develop a forecasting model using a guided forward step-wise regression approach. They found that time-of-day, temperature and humidity are significant predictors. Thomas et al. \cite{thomas2009temporal} studied cycle flows from utilitarian and recreational paths in the Netherlands. A bi-level model for predicting the demand for cycling was used. The lower level describes how cyclists value the weather. The upper level is the relation between demand and this weather value. Most fluctuations are described by the model. 

In the third category, studies that have used context to enrich forecasting using neural networks, Thu et al. \cite{thu2017multi} propose multi-layer perceptron regressors from multi-source context data to predict bike pick-up demands from New York city BSS considering clusters of stations based on their geographical locations and transition patterns. The proposed networks combine weather factors (condition, temperature, wind speed, and visibility) and taxi trip records. Despite its relevance, temporal dependencies between observations are disregarded. 
Pan et al. \cite{pan2019predicting} incorporate weather record data at input layer of LSTMs to improve the prediction of bike sharing demand for balancing of distribution of bikes across stations. Results evidenced improvements against context-unaware LSTMs. Recent contributions on deep learning research also show the possibility of incorporating specific forms of calendrical and spatial awareness \cite{pan2019urban,casas2019spatially}.
\section{Solution: context-aware LSTMs}
\label{sol-sec}

We propose a new recurrent neural network layering able to incorporate both historical and prospective sources of context to guide forecasting tasks. The proposed architecture is a sequential composition of two components, $C_1$ and $C_2$. $C_1$ is a LSTM that receives context-enriched multivariate inputs and returns the forecasted series as the output. 
$C_2$ is a LSTM (or a gated recurrent unit) that receives as input the forecasted series from $C_1$ and prospective sources of context along the horizon of prediction and returns the true forecasting. Sections \ref{solmasking} and \ref{solprospective} respectively describe how historical and prospective sources can be fed into the proposed architecture, and further discuss their significance. 

\subsection{Incorporating historical context: masking}
\label{solmasking}

To incorporate historical sources of context data, we take advantage of the fact that LSTMs are inherently prepared to learn mapping functions from multivariate time series into the target univariate time series (demand variable along a prediction horizon). In this way, an arbitrarily-high multiplicity of context variables can be combined at the input layer of the $C_1$ component 
to guide the learning task. 
Sections \ref{solcalendrical} to \ref{solspatial} offer important masking principles on how to compose the multivariate time series from difference context sources. 
For the purpose of illustrating the principles introduced along these sections, consider that the task at hands is to learn a predictive model that forecasts the number of hourly bicycle check-outs for the upcoming days at a given station using a two-year historical data with an hourly aggregation, 

\textcolor{white}{.}\vskip -0.9cm
\begin{align*}
    \vec{x}_{\text{example}} &= (\vec{x}_{0h:1/1/2018}, \vec{x}_{1h:1/1/2018}, .., \vec{x}_{23h:31/12/2019}),\\
    &= ((9), (6), .., (10)),
\end{align*}

\noindent a series to be further segmented for the purpose of learning. 

Under masking principles, sources of historical context are hypothesized to guide the forecasting task in two major ways. First, by looking to the past and modeling how a certain context variable affects bike demand at a given station offers the possibility of removing context-dependent factors. Considering wind intensity as an illustrative context variable, deviations from regular wind ranges can be considered by the LSTM unit to better model the co-observed demand. 

Second, some sources of historical context can be correlated with future context and/or demand. Therefore, this relation can be directly learned by the LSTM unit to shape short-term forecasts. For instance, strong wind intensity at a given hour may decrease the observed demand with effect lasting up to three hours.

\subsubsection{Station context}
\label{solstation}
In addition to the target variable, complementary demand variables 
can be inputted to guide the prediction. For instance, the analysis of check-ins can be complemented with the volume of check-outs to account for correlated deviations. 

Another example is to consider the load state of the station as we know that states of high-load (low-load) can impact check-outs (check-ins). 
Illustrating, $\vec{x}_{\text{example}}$=((\textit{check-in}=9,\textit{check-out}=6,\textit{load}=0.9), ..,(\textit{check-in}=10,\textit{check-out}=12,\textit{load}=0.3)), is an augmented series with a multivariate order of 3.

\subsubsection{Calendrical context}
\label{solcalendrical}

Four masks are suggested:
\begin{itemize}
    \item a \textit{day mask} can be added to the time series with information pertaining to the week-day of each observation,
    \begin{equation*}
        \vec{x}_{\text{example}} = ((9,monday),(6,monday),..,(10,tuesday)).
    \end{equation*}
    In this way, LSTMs are inputted with information that helps them internally separating neural pathways in accordance with the week-day. This is a simplistic yet essential guiding information whenever segmentation creates instances starting at different timings. 
    
    In addition, this masking principle can act as way of augmenting data as it allows for the combined inclusion of weekends and weekday periods in the forecasting task, instead of learning separate forecasting models. 
    
    If weekdays show similar patterns of demand, simpler daily masks can be applied using mappings of lower cardinality, such as \textsf{\{weekday, saturday, sunday\}};
    
    \item a \textit{holiday mask} can be added as a separated binary variable in order to indicate whether a given observation was produced or not within a holiday;
    \item an \textit{hour mask} can be added as an additional variable to guide the learning task whenever the segmentation step does not guarantee that every data instance starts at the same time step. As a result, daily seasonalities can be implicitly captured within the LSTM as a non-linear function of the hour within the day. If the time granularity of the demand series is finer (coarser) than hour, a finer (coarser) mask can be produced. For instance, a mapping \textsf{\{dawn, morning, afternoon, evening\}} is adequate for temporally misaligned data instances and time windows with up to 6 hours;
    \item other relevant masks, include \textit{academic period masks} or \textit{festivity period masks} that may not necessarily coincide with the aforementioned holiday mask. In the context of our work, the influence of these masks were found to be essential.
\end{itemize}

\subsubsection{Situational context}
\label{solsituational}

Similarly to the masks produced from year and academic calendars, situational masks mark periods where events of interest may impact the demand observed at a given station. The major difference between these sources of context is the fact that situational context is usually circumscribed to a specific geographical area, therefore only impacting a subset of stations from a given BSS. Given a station, the situational variables produced using these masking principles generally correspond to public events (concerts, conventions or sport events) that occur within a maximum distance from the target station.

Situated events can produce varying levels of impact on the demand at a given station. For instance, stations nearby a stadium and a small concert hall should be able to differentiate between sport events and concerts. In this context, these masks can explicitly capture the magnitude of an event, as well as variations on that same magnitude throughout the period where the effects of the event lasts. Illustrating,  $(0,0,1,3,3,1,1,3,3,1,0...)$ series may correspond to a mask modeling the magnitude of a gathering that can be placed as a complementary inputted series to guide the learning.

\subsubsection{Meteorological context}
\label{solmeteorological}

Meteorological variables, including wind intensity and precipitation, can be as well inputted to guide the learning. The closest meteorological station can be selected or k-nearest stations to compute the weather series (under distance-weighting schema). Additional weather variables can be produced from the raw weather records produced by meteorological stations, including the perceived temperature or integrative scores. 

A high number of weather variables leads to high a multivariate order of the inputted, which can hamper the learning if not enough historical observations are available. In this context, variable selection can be considered to guarantee that only the most informative weather variables are inputted.

\subsubsection{Spatial context}
\label{solspatial}

In the context of station-wise dependencies, we propose a geographical mask, referred as the nearby mask. The nearby mask is the occupation ratio of the neighbouring stations. The algorithm to create the nearby variables, Algorithm \ref{algorithm:create_mask_nearby}, receives the target station or cluster of stations under analysis, $SS$, a distance radius, $R$, list of all stations, $S$, and time series $TS$. Algorithm \ref{algorithm:create_mask_nearby} has three-steps: 1) computes the centroid of the selected stations, line 3; 2) finds all the stations inside the $R$ radius using the previous centroid as the centre (lines 4--8), and updates the time series with the occupation ratio of the nearby stations (lines 10--12).

Under the nearby mask, the target LSTMs are now able to consider the impact that stations with high- and low-load have on the demand of nearby stations. 
Here, the occupation ratio is proposed in contrast with full-and-empty station states to allow for more flexible learning (nearly full stations may be full in the upcoming time steps). Still, alternative encodings can be proposed by customizing line 11 of Algorithm \ref{algorithm:create_mask_nearby}.

\vskip -0.25cm
{\small
\begin{algorithm}[!h]
    
    \SetKwFunction{FMain}{create\_nearby\_mask}
    \SetKwProg{Fn}{Function}{:}{}
     \Fn{\FMain{$SS, S, R, TS$}}{
        nearby\_station = []
        
        lat\_center, long\_center = calculate\_centroid($SS$)
        
        \ForEach{station $s_i \in S$}{%
        
            \mbox{distance = Euclidean((lat\_center, long\_center), ($s_i$.lati, $s_i$.long))}
            
            \If{distance $<$ radius}{%
                nearby\_station.append($s_i$)
            }
            
        }
        
        \ForEach{station $s_i \in $ nearby\_station}{
        
        $TS$ = update\_series \Big($TS,\dfrac{s_i._{\text{num\_bikes}}}{s_i._{\text{location\_capacity}}}$\Big)
        
        }
    
    }
\textbf{End Function}

    \caption{Create nearby mask algorithm}
    \label{algorithm:create_mask_nearby}
\end{algorithm}
}

\subsection{Incorporating prospective context: time-dependent regularization}
\label{solprospective}

Prospective sources of context, including weather forecasts or planned events, may be available along the horizon of prediction. In this context, their use can further guide the learning. Recall that the proposed architecture is a composition of two components, $C_1$ and $C_2$. $C_1$ is a LSTM that receives multivariate inputs (in accordance with the principles introduced along section \ref{solmasking}) and returns the forecasted series as the output. $C_2$ is a recurrent neural component that receives the forecasted series from $C_1$ and the prospective sources of context along the horizon of prediction as inputs, and returns the true forecasting. In this context, $C_2$ can be though as a context-aware denoiser or time-dependent regularizer of the forecasts. 
Empirical analysis show optimal performance when $C_2$ is a LSTM unit. Nevertheless, a gated recurrent unit (GRU) provides a competitive alternative given the properties of this 
step. 

Under the proposed architecture, masking principles -- similarly to the ones introduced along section \ref{solmasking} -- can be considered to compose the multivariate time series to be inputted to the $C_2$ component. In this context:
\begin{itemize}
    \item calendrical masks can be produced to regularize the forecasts whenever the prediction horizon contains holidays or any other meaningful season (such as academic break). In addition, and more generally, time and day masks can be as well inputted for a calendric-guided revision of forecasts;
    \item situational masks can be produced from planned events available from cultural and sport agendas, as well as semi-structured repositories containing information on the usage of public spaces. 
    See section \ref{solsituational} for principles on duration- and intensity-sensitive encodings for situational masks;
    \item meteorological masks can be produced from weather forecasts. As weather forecasts are typically produced under a coarse-grained time granularity, upsampling may be necessary to guarantee the weather and demand series share the same granularity. See section \ref{solmeteorological} for principles on station and variable selection;
    \item spatial masks can be derived from forecasts of nearby stations' load (see principles in section \ref{solspatial}) when load forecasts are able to satisfy upper bounds on the observed error. 
\end{itemize}

By incorporating prospective sources of context, 
the $C_1$-forecasted demand series can now be further subjected to context-dependent corrections, guiding the forecasting task. For instance, periods with forecasts of high wind intensity can suffer an adjustment on the $C_1$-based demand expectations by capturing relationships between the multivariate input series and the true forecast. 

\subsection{Learning schema}
\label{sol:schema}

To guarantee the proper learning of the proposed neural networks, considerations associated with both the segmentation and the learning setting should be placed. Given a specific horizon of prediction $h$ and its time resolution $\Delta$, the series should be segmented in a set of instances, each instance being a pair (\textit{input}, \textit{output}) where the \textit{input} is a series with at least twice the length of the prediction horizon (by default $7\times h$), the \textit{output} is the subsequent $h$-length series, and both series have the same time resolution, $\Delta$. Segmentation is further characterized by the presence of a sliding window to compose the data instances. By default, the sliding window corresponds to a single day (24 hours). 

\begin{figure}[H]
\vskip -0.2cm
    \centering
        \includegraphics[scale=0.35]{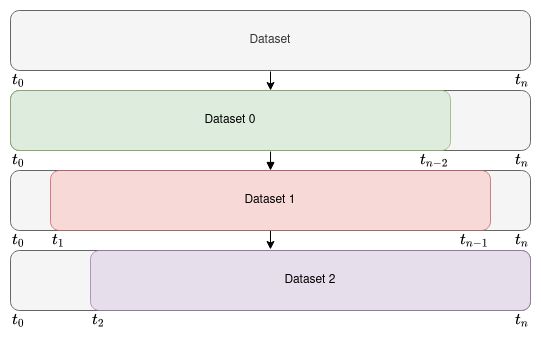}
\vskip -0.3cm
    \caption{\small Example of algorithm \ref{algorithm:create_sub_dataset} with $step\_size$ of 1}
\vskip -0.2cm
    \label{fig:example_training_methodology}
\end{figure}

The above principles are applied to segment series in the context of a single dataset. For a more robust validation of predictors, cross-validation is suggested. Particular attention should be paid to guarantee the soundness of the cross-validation setting given the inherent time frame of the series data. Algorithm \ref{algorithm:create_sub_dataset} shows how multiple datasets are produced. In this context, \textit{series} is the inputted (univariate or multivariate) time series with a time frame $\Delta$ given by the user; \textit{window\_size} is the length in days of each dataset, and \textit{step\_size} is the lag value between datasets in days. 
Figure \ref{fig:example_training_methodology} provides an example of the algorithm \ref{algorithm:create_sub_dataset}, where \textit{window\_size} = $T-2$, \textit{step\_size} = 1 and \textit{datasets} is the resulting set of three datasets.
\vskip 0.15cm
\begin{algorithm}[!ht]
\small
    \SetKwFunction{FMain}{create\_subdatasets}
    \SetKwProg{Fn}{Function}{:}{}
     \Fn{\FMain{dataset, window\_size, step\_size}}{
     datasets = \{\}
        
        end\_dataset\_index = window\_size - 1 
        
        \While{\textit{end\_subdataset\_index $\le$ $|$series$|$}}{
            
            \mbox{begin\_dataset\_index = end\_dataset\_index - window\_size}
            
            \mbox{dataset = series[begin\_dataset\_index:end\_dataset\_index]}
            
            datasets =  datasets $\cup$ \{dataset\}
            
            end\_dataset\_index += step\_size
        }
              \textbf{return} datasets
     }
    \textbf{End Function}

    \caption{Split dataset algorithm }
    \label{algorithm:create_sub_dataset}
\end{algorithm}
\vskip 0.15cm

After applying Algorithm \ref{algorithm:create_sub_dataset}, we split each sub-dataset into train, validation and test data, guaranteeing that training data instances always precede validation and testing data instances in order to better mimic real-time testing scenarios. 

\subsubsection{Hyperparameter otimization}
\label{sol:hyperparam}
Each dataset is scaled using a min-max normalization to improve the LSTM performance. The min-max scaler parameters are learned from the training data instances for each dataset. The test and validation segments are also normalized using the scaler learned from the training segment. 

Parameterizations associated with the proposed layering architecture were empirically conducted using a sensitivity analysis on the number, size, type and composition of layers, as well as on the results yield by different loss functions (including cosine, mean absolute error, mean squared error) and forms of regularization. 

All the additional parameters -- including activation functions, presence of drop-out layers, optimizers, batch size and learning rates -- were fixed using Bayesian optimization. The optimized metric was the mean absolute error produced over all the training segments. Early stopping from validation errors is considered.

\section{Results}
\label{res-sec}

Results are organized in three major steps. First, we introduced background considerations on the: i) target public BSS (GIRA) and considered sources of context, ii) assessment setting, and iii) general exploratory statistics on GIRA's bike demand. Second, we provide a thourough comparison of state-of-the-art forecasters against the proposed context-aware predictors. Third, we offer a greater detail on the performance of the proposed context-aware predictors, describing the impact that different sources of context have on the forecasting residues. 

The proposed context-aware predictors were implemented using \textsf{tensorflow}, \textsf{python}. The remaining state-of-the-art predictors were implemented using facilities from \textsf{statsmodels}, \textsf{tslearn} (barycenter calculus) and \textsf{scikit-learn} packages in \textsf{python}. The parameters of the state-of-ther-art predictors (such as smoothing factors in Holt-Winters or $k$ in $k$-nearest neighbor regressors) were subjected to Bayesian optimization.
\vskip -0.5cm\textcolor{white}{.}

\subsection*{Case study: GIRA} \vskip -0.1cm
To comprehensively assess the proposed contributions, we consider the Lisbon's public BSS, termed GIRA, under the joint responsibility of EMEL and the Lisbon's city Council. 
The data associated with the target GIRA network contains all bike trip records from November 2018 until April 2019, with timestamps for every change in stations' state and the corresponding load value. Structural changes to the stations' capacity associated with the BSS expansion are also considered. Given the station load data, the check-out and check-in events within the Lisbon's public BSS were inferred from the differences on the load state of each station along time. 
\vskip -0.5cm\textcolor{white}{.}

\subsection*{Lisbon's urban context}\vskip -0.1cm
Historical and prospective \textit{weather record data} was sourced by Instituto Português do Mar e da Atmosfera (IPMA) and Instituto Superior Técnico (IST). Humidity, wind speed (km/hour), pressure, precipitation  and temperature (Celsius) weather variables were sourced. Four meteorology stations available. The station closer to a station is selected for providing weather data. When analyzing clusters of stations, the centroid of each centroid is computed in order to identify the closest meteorology station. 

\textit{Situational context data} was derived from: the Lisbon's cultural agenda, road construction works, and the city Council's records on the occupation of large public spaces and stadiums. Finally, \textit{calendrical context data} was sourced from available civil and academic calendars, and \textit{spatial context data} inferred from the GIRA dataset.

\vskip -0.5cm\textcolor{white}{.}

\subsection*{Station clusters}\vskip -0.1cm

\begin{figure}[!ht]
\vskip -0.15cm
    \centering
         \begin{subfigure}[b]{0.15\textwidth}
            \centering
            \includegraphics[width=\textwidth]{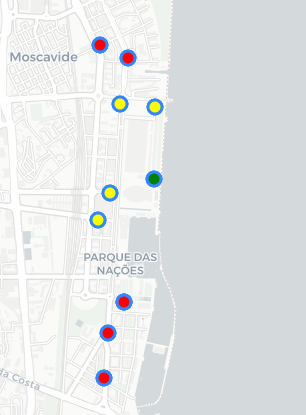}
            \caption{\small Oriente cluster}
            \label{fig:oriente_cluster}
        \end{subfigure}
         \begin{subfigure}[b]{0.15\textwidth}
            \centering
            \includegraphics[width=\textwidth]{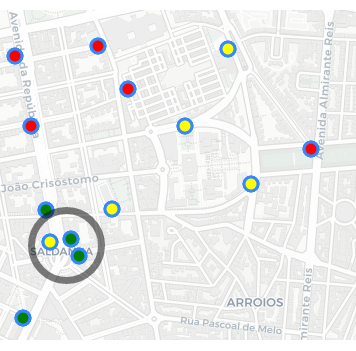}
            \caption{\small Saldanha cluster}
            \label{fig:saldanha_cluster}
        \end{subfigure}
        \begin{subfigure}[b]{0.15\textwidth}
            \centering
            \includegraphics[width=\textwidth]{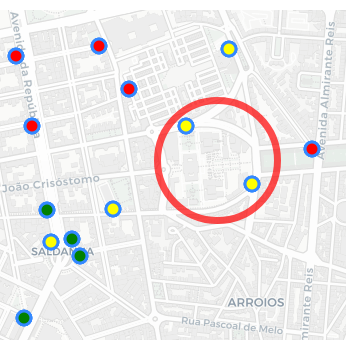}
            \caption{\small IST cluster}
            \label{fig:ist_cluster}
        \end{subfigure}
\vskip -0.2cm
\caption{\small Selected clusters of stations (screenshot from ILU App)}
\vskip -0.2cm
    \label{fig:clusters_map}
\end{figure}

Bike demand analysis can be pursued at different spatial granularities in accordance with the targeted aim, such as establishing balancing initiatives (finer granularity) or general planning decisions (coarser granularity). To this end, we predict bike demand at three distinct levels: i) station level; ii) cluster of stations; and iii) BSS level (all stations). 
To help the city Council understanding demand, we identify three clusters of interest: \textsf{IST} cluster, \textsf{Saldanha} cluster and \textsf{Oriente} cluster, represented in Figure \ref{fig:clusters_map}. 

{}

Each of these clusters yield unique aspects of interest. The \textsf{IST} cluster has a high share of students since its stations are close to the campus of Instituto Superior Técnico's faculty. \textsf{Saldanha} cluster has a high share of workers since its stations are located within a well-known business district. Finally, \textsf{Oriente} cluster has a high share of tourists since its stations are located near Parque das Nações, a place with diverse cultural, expo and leisure attractions. 

The \textsf{Oriente} cluster is the larger cluster, encompassing 6 docking stations and a total capacity of 144 bicycles. The \textsf{Saldanha} cluster has 3 stations with a total capacity of 44 bicycles. Even though, \textsf{IST} cluster has only 2 stations, it is capable of holding 51 bicycles.
\vskip -0.4cm\textcolor{white}{.}

\subsection*{Exploratory statistics}
\begin{figure}[!ht]
\vskip -0.3cm
    \centering
        \includegraphics[width=.98\columnwidth]{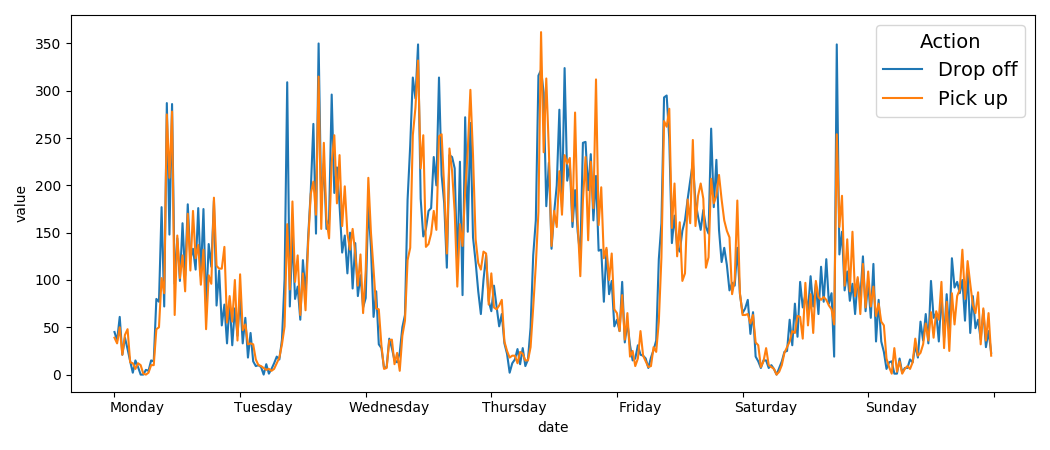}
\vskip -0.2cm
    \caption{\small Half-hourly volume of check-outs/ins within GIRA.} 
\vskip -0.2cm
    \label{fig:linechart_check_in_checkout}
\end{figure}

\begin{figure}[!ht]
\vskip -0.3cm
    \centering
        \includegraphics[width=1\columnwidth]{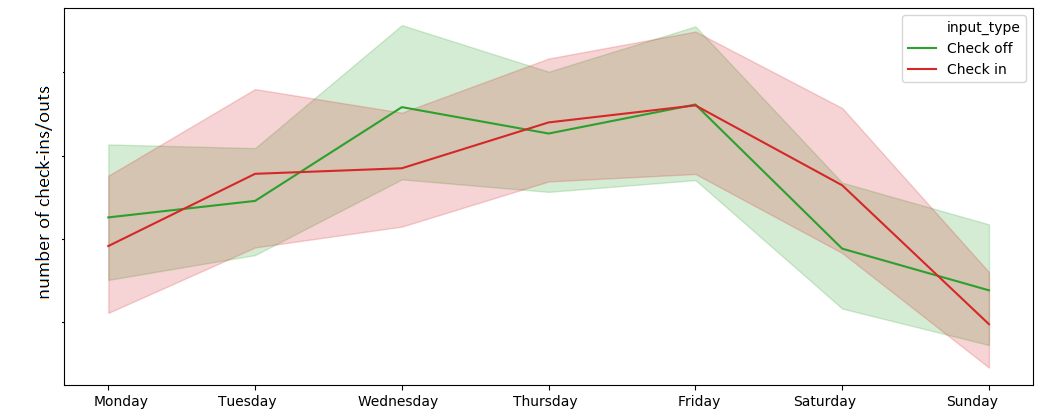}
\vskip -0.2cm
    \caption{\small Daily volume and variation of check-outs and check-ins at a single GIRA's station (November 2018 to April 2019).}
\vskip -0.3cm
    \label{fig:linechart_check_in_checkout2}
\end{figure}

Figures \ref{fig:linechart_check_in_checkout} and \ref{fig:linechart_check_in_checkout2} provide a high-level view on how the demand varies for the GIRA's BSS as whole (cumulative demand from all stations). Figure \ref{fig:linechart_check_in_checkout} offers a view on how demand is distributed along a normal week, evidencing the calendrical differences and the effect produced by public events at Saturday night. Figure \ref{fig:linechart_check_in_checkout2} captures the variability of demand along the targeted period, where the bounds (standard deviation) confirm that bike sharing demand is susceptible to diverse idiosyncrasies causing significant variability.

{}

{}

\subsection*{Comparison of state-of-the-art forecasters}

Tables \ref{tab:commands1} and \ref{tab:commands2} gather the predictive performance of state-of-the-art forecasters, focusing respectively on check-ins and check-outs at different spatial granularities (system-, cluster- and station-wise levels). The prediction horizon is 24 hours under a 30-minutes time step. The mean absolute error (MAE) and root mean squared error (RMSE) were the selected forecasting metrics. The average and standard deviation estimates for these metrics were computed from the residues gathered from all testing data instances from each one of the three data folds (according to section \ref{sol:schema}). 

From the state-of-the-art predictors, a variant of existing kNN forecasters, combining neighborhood search and barycenter computation from sets of $k$ time series, is proposed given its superior performance. In this  context, barycenter forecasters can be though as kNN forecasters when $k$ equals the number of training data instances.  Dynamic time wrapping (DTW) is considered as a similarity metric to tolerate temporal misalignments.

Five major remarks are observed. First, classical approaches, including Holt-Winters, are unable to properly deal with the arbitrarily-high variability of bike sharing demand as seasonal and trend modeling is generally insufficient to handle the peculiarities of the targeted bike trip data. Second, distance-based predictors yield slightly worse yet competitive results against the proposed LSTM-based predictors. Third, the proposed context-aware LSTM-based predictors generally offer slight improvements against context-unaware LSTM-based predictors, yet the improvements do not yield statistically significant differences\footnote{Residues normally distributed. Unilateral paired $t$-tests with $p$-value above 0.01.}. Fourth, from the different sources of context, the individual use of meteorological sources and the joint use of meteorological-and-spatial sources of context are the optimal settings. Finally, historical and prospective sources of context have similar impact on the forecasting performance.

\begin{figure}[!ht]
    \centering
\vskip -0.2cm
    \includegraphics[width=0.5\textwidth]{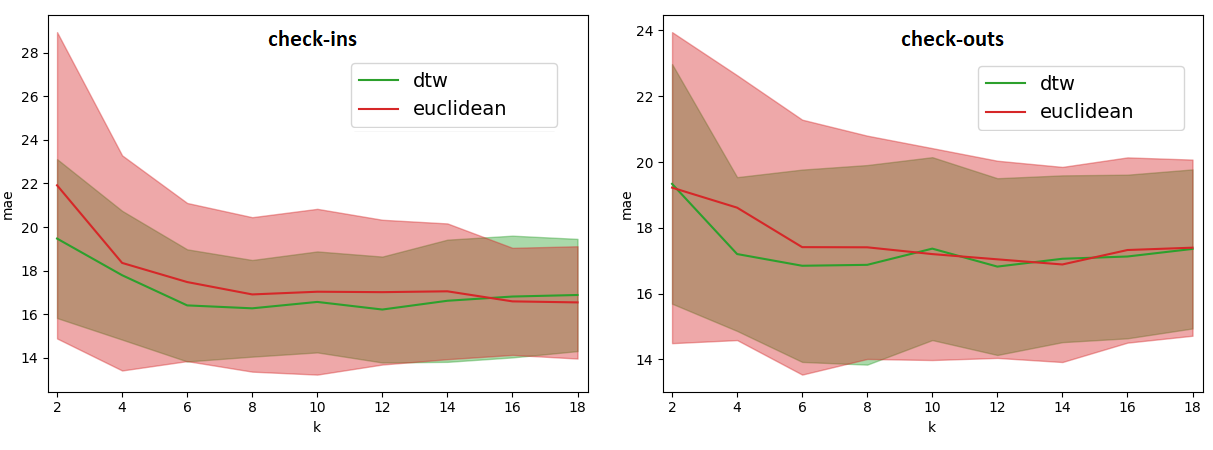}
            \vskip -0.15cm
    \caption{\small MAE performance of proposed $k$NN with varying $k$ and distance (Euclidean vs DTW in neighborhoods and barycenters) to forecast demand for the whole GIRA network.}
    \vskip -0.15cm
    \label{fig:multiple_k}
\end{figure}

Figure \ref{fig:multiple_k} provide a more detailed view on the performance of the proposed $k$NN predictors, showing the impact of locality -- $k\ge 5$ --, and tolerating misalignments -- DTW is comparable to Euclidean distance to compute both neighborhoods and barycenters.
\vskip -0.2cm

{}

\subsection*{Context impact on demand predictors}
\begin{figure}[!ht]
    \vskip -0.1cm

        \includegraphics[width=0.81\columnwidth]{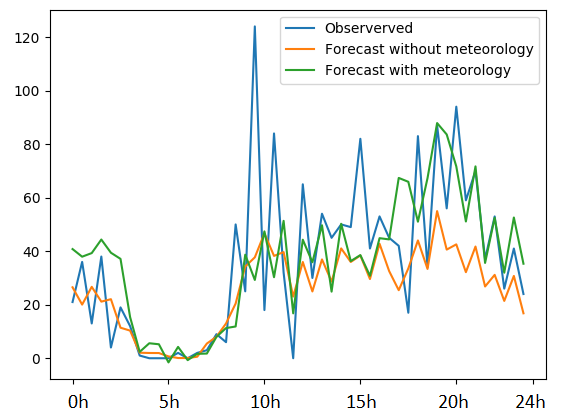}
    \vskip -0.25cm
    \caption{\small Aleatory testing data instance: true observations vs. LSTM forecasts vs. weather-aware LSTM-based forecasts considering check-ins for all GIRA stations.}     \vskip -0.2cm
    \label{fig:all_forecast_meteo_vs_forecast_no_meteo}
\end{figure}

\begin{figure}[!ht]
    \vskip -0.15cm
    \centering
        \includegraphics[width=0.76\columnwidth]{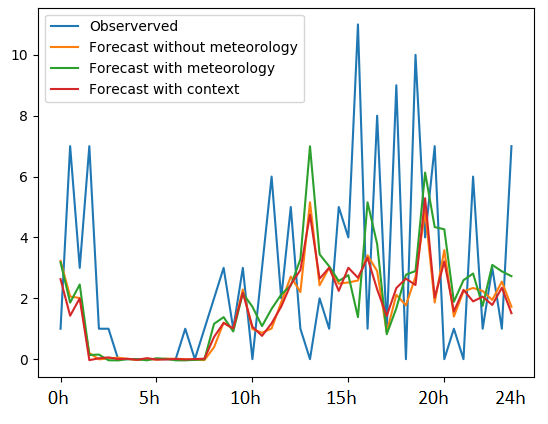}
    \vskip -0.25cm
    \caption{\small Hard testing data instance: true observations vs. LSTM forecasts vs. context-aware LSTM-based forecasts considering check-ins for \textsf{Oriente} cluster of stations.} 
        \vskip -0.15cm
\label{fig:oriente_forecast_meteo_vs_forecast_no_meteo_context}
\end{figure}

Figures \ref{fig:all_forecast_meteo_vs_forecast_no_meteo} and \ref{fig:oriente_forecast_meteo_vs_forecast_no_meteo_context} provide two illustrative data instances that clearly show the impact of incorporating meteorological context data on the forecasts. Figure \ref{fig:all_forecast_meteo_vs_forecast_no_meteo} considers the check-in demand at the whole network level, clearly showing the indisputable role of incorporating weather record data for improving predictions. Figure \ref{fig:oriente_forecast_meteo_vs_forecast_no_meteo_context} considers the check-in demand at the \textsf{Oriente} cluster level, showing the impact from the integration of multiple sources of context.

\begin{table*}[!t]
\footnotesize
  \caption{\small Comparison of forecasting errors (MAE and RMSE) of state-of-the-art predictors: \textit{check-in demand} at 30 minutes granularity.}
  \label{tab:commands1}
  \vskip -0.26cm
  \begin{tabular}{lllllll}
      \toprule
   & \multicolumn{2}{l}{Senario 1: all stations} & \multicolumn{2}{l}{Senario 2: \textsf{Oriente} cluster} & \multicolumn{2}{l}{Senario 3: single station} \\
       \midrule

   & MAE & RMSE & MAE & RMSE & MAE & RMSE \\
       \midrule
    
KNN - DTW  & 16.57$\pm$ 2.27  & 23.52$\pm$ 3.84 & 2.42$\pm$0.23 &3.81$\pm$0.48 & 0.35$\pm$0.06 & 0.79$\pm$0.14 \\
KNN - Euclidean & 17.03$\pm$3.72 & 24.06$\pm$6.02 &2.46$\pm$0.35 &3.82 $\pm$0.64 &  0.34$\pm$0.06 &0.78 $\pm$0.14\\
       \midrule
    Barycenter - DBA & 19.81$\pm$3.32 & 27.87$\pm$4.86 &2.47$\pm$0.43 &3.82 $\pm$0.54 & 0.34 $\pm$ 0.05 & 0.85 $\pm$ 0.15\\
Barycenter - Euclidean & 17.01$\pm$3.42 & 24.81$\pm$5.35 &2.11$\pm$0.25 &3.25 $\pm$0.44 & 0.35 $\pm$0.05 &0.70 $\pm$ 0.21\\
    Barycenter - DTW & 18.57$\pm$3.28 & 26.87$\pm$4.92 &2.16$\pm$0.26 &3.35 $\pm$0.46 & 0.36 $\pm$ 0.05 & 0.71 $\pm$ 0.22\\
       \midrule
Holts-Winters & 24.11$\pm$6.19 & 33.94$\pm$14.88 &3.17$\pm$0.45 &4.64 $\pm$0.45& 1.01$\pm$ 0.33& 1.99 $\pm$ 0.73\\
       \midrule
    LSTM check-in & 15.07$\pm$3.41 &  20.87$\pm$3.4 &2.26$\pm$0.23 &3.45 $\pm$0.23&  0.24$\pm$ 0.05& 0.72 $\pm$ 0.14\\
LSTM check-in and check-out& 14.49$\pm$3.14 & 20.48$\pm$3.15 &2.17$\pm$0.22 &3.38 $\pm$0.21& 0.24 $\pm$ 0.05 & 0.72 $\pm$ 0.14\\
 LSTM day mask & 13.97$\pm$3.97 & 20.35$\pm$3.97 & 2.24$\pm$0.23 &3.51 $\pm$0.22 & 0.24$\pm$ 0.05& 0.72 $\pm$ 0.14\\
  LSTM nearby & - & - &2.23$\pm$0.26 &3.38 $\pm$0.26 & 0.36 $\pm$ 0.05 & 0.68 $\pm$ 0.14\\
 LSTM all meteo & 14.75$\pm$3.80 & 21.01$\pm$3.80 & 2.31$\pm$0.19 &3.62 $\pm$0.19 & 0.24 $\pm$ 0.05& 0.72$\pm$0.14\\
 LSTM all meteo and nearby & - & - & 2.12$\pm$0.27 &3.45 $\pm$0.68 & 0.36 $\pm$ 0.05& 0.68 $\pm$ 0.14\\
       \bottomrule
  \end{tabular}
\end{table*}

\begin{table*}[!t]
\footnotesize
  \caption{\small Comparison of forecasting errors (MAE and RMSE) of state-of-the-art predictors: \textit{check-out demand} at 30 minutes granularity.}
  \vskip -0.26cm
  \label{tab:commands2}
  \begin{tabular}{lllllll}
      \toprule
   & \multicolumn{2}{l}{Senario 1: all stations} & \multicolumn{2}{l}{Senario 2: \textsf{Oriente} cluster} & \multicolumn{2}{l}{Senario 3: single station} \\
       \midrule

   & MAE & RMSE & MAE & RMSE & MAE & RMSE \\
    \midrule
    
KNN - DTW  & 17.37$\pm$ 2.72  & 24.21$\pm$3.64  & 2.44$\pm$0.42 &3.64$\pm$0.55 & 0.36$\pm$0.05 & 0.88$\pm$0.18 \\
KNN - Euclidean & 17.20$\pm$3.15 & 23.78$\pm$4.43 &2.54$\pm$0.33 &3.77 $\pm$0.44 & 0.34$\pm$0.06 &0.81 $\pm$0.21\\
\midrule
Barycenter - DBA & 19.70$\pm$3.06 & 27.18$\pm$5.05 &2.35$\pm$0.26 &3.82 $\pm$0.56 & 0.36 $\pm$ 0.08 & 0.81 $\pm$ 0.14\\
Barycenter - Euclidean & 15.62$\pm$3.18 & 22.30$\pm$5.29 &2.11$\pm$0.29 &3.33 $\pm$0.55 & 0.36 $\pm$0.05 &0.67 $\pm$ 0.14\\
    Barycenter - DTW & 18.25$\pm$2.95 & 25.85$\pm$4.96 &2.16$\pm$0.31 &3.48 $\pm$0.56 & 0.36 $\pm$ 0.03 & 0.66 $\pm$ 0.12\\
    \midrule
Holts-Winters & 25.00$\pm$4.94 & 34.78$\pm$4.94 &2.91$\pm$0.35 &4.21 $\pm$0.35& 3.10$\pm$ 1.48& 4.14 $\pm$ 1.79  \\
    \midrule
LSTM check-out & 15.73$\pm$1.59 & 22.31$\pm$3.59 &2.32$\pm$0.29 &3.37 $\pm$0.29x&  0.35 $\pm$ 0.05& 0.71 $\pm$ 0.21\\
LSTM check-in and check-out& 16.08$\pm$1.84 & 23.00$\pm$3.22 &2.24$\pm$0.22 &3.46$\pm$0.41& 0.35 $\pm$0.05 & 0.70$\pm$ 0.21\\
 LSTM day mask & 16.60$\pm$4.50 & 23.35$\pm$4.50 & 2.34$\pm$0.32 &3.48 $\pm$0.32 & 0.36 $\pm$ 0.05 & 0.71 $\pm$ 0.21\\
  LSTM nearby & - & - &2.16$\pm$0.22 &3.30 $\pm$0.44 & 0.35$\pm$ 0.05 & 0.71 $\pm$ 0.21\\
  LSTM all meteo & 16.02$\pm$1.70 &23.14$\pm$3.40 &2.42 $\pm$0.32 & 3.62$\pm$ 0.49 & 0.36 $\pm$ 0.06 & 0.72$\pm$ 0.23\\
  LSTM all meteo and nearby & - & - &2.16 $\pm$0.22 & 3.30$\pm$ 0.43 & 0.35 $\pm$ 0.05 & 0.71$\pm$ 0.21\\
         \bottomrule

  \end{tabular}
\end{table*}

{}

{}

\begin{figure}[!b]
    \vskip -0.15cm
    \centering
        \includegraphics[width=0.68\columnwidth]{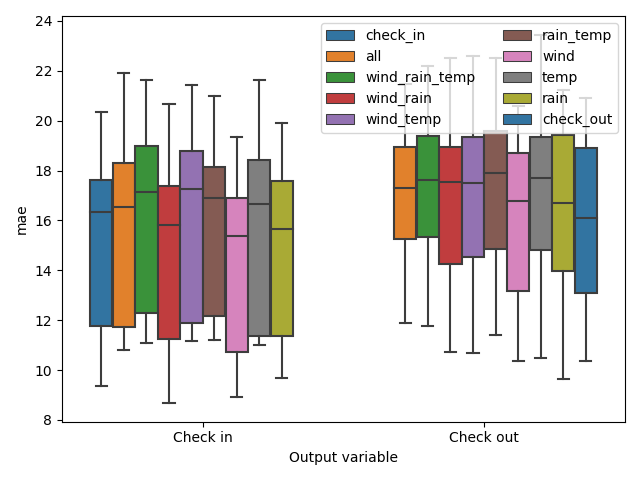}
    \vskip -0.25cm
    \caption{\small Performance impact of historical and prospective weather variables on the forecasted check-ins at GIRA's network.} 
    \vskip -0.15cm
    \label{fig:compare_meteo}
\end{figure}{}

\begin{figure}[!ht]
    \centering
    \vskip -0.15cm
        
        \includegraphics[width=1\columnwidth]{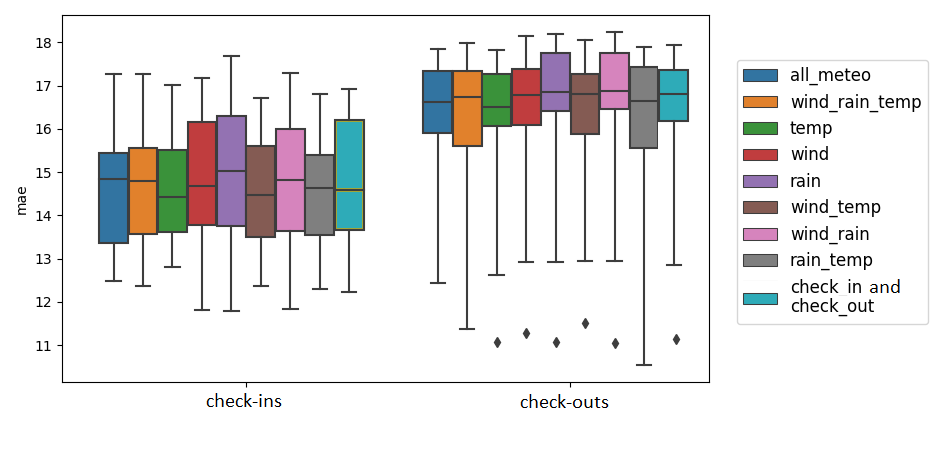}
          \vskip -0.15cm

    \caption{\small Performance impact of joint \textit{prospective-and-historical context} variables on the demand forecasts at GIRA's network.}
        \vskip -0.15cm
\label{fig:compare_meteo_historical_forecast}
\end{figure}

Figures \ref{fig:compare_meteo} and \ref{fig:compare_meteo_historical_forecast} measure the impact from incorporating different context variables, including historical and available prospective variables. Previous experimental settings are preserved for this analysis. Three major observations. First, from weather variables, wind intensity and precipitation variables offer the largest guidance. Second, the use of complementary demand series (section \ref{solstation}) and spatial masks (section \ref{solspatial}) impact predictions as well, although the differences on the averaged error are not statistically significant. Finally, we observe that the incorporation of historical context has slightly higher positive impact on predictions in comparison with prospective context. We hypothesize that this observation is a result of the inherent uncertainty factors associated with the sources of prospective context.

{}

\subsection*{Context-aware forecasters: parameters}

\begin{figure}[!ht]
\vskip -0.2cm
    \centering
\includegraphics[width=.5\textwidth]{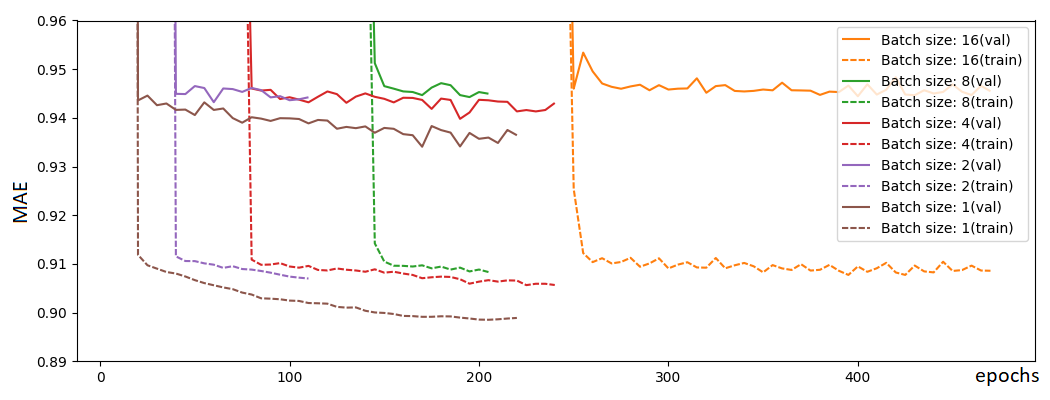}
            \vskip -0.2cm
    \caption{\small \mbox{Batch size impact on GIRA network check-ins prediction.}}
    \label{fig:multiple_bacth_size}
\end{figure}

\begin{figure}[!ht]
    \centering
    \vskip -0.35cm    \includegraphics[width=.47\textwidth]{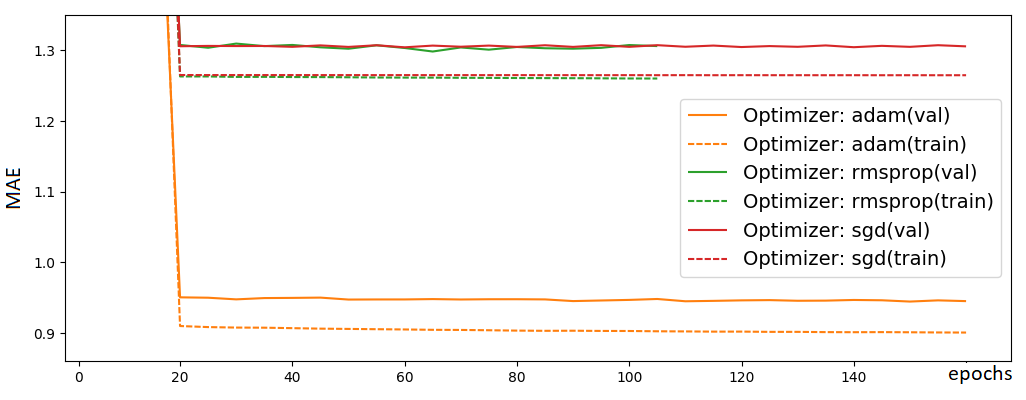}
            \vskip -0.2cm

    \caption{\small \mbox{Optimizer impact on check-out forecasts at GIRA network.}}
    \label{fig:multiple_optimizer}
            \vskip -0.2cm
\end{figure}

Figures \ref{fig:multiple_bacth_size} and \ref{fig:multiple_optimizer} provide two illustrative analyzes of the impact of two parameters -- batch size and optimizer -- on the performance of the target context-aware predictors. These figures suggest the use of small batch sizes and Adam optimization. Similar analyzes conducted for the parameters listed in section \ref{sol:hyperparam} confirm the adequacy of the fixed architectural and learning parameters. Additional illustrative parameters include the presence of drop-out layers, $1E-5$ learning rate, and $L_1$ norm regularization. 

{}

{}

\section{Discussion}
\label{disc-sec}

This work proposed neural processing principles to leverage the performance of prediction tasks in the presence of heterogeneous sources of context data. Motivated by the intrinsic challenges associated with the operation and planning of bike sharing systems, we show the relevance of the proposed approach for demand analysis in this domain, particularly hampered by the spatial dependencies between stations, amongst other significant externalities.

Two major contributions were introduced to this end. First, we propose a serial composition of LSTMs, taking advantage of their inherent ability to capture cross-variable relationships from multivariate time series data. In this context, we propose masking principles to derive complementary series' variables from spatial, meteorological and calendrical sources of context. Second, we propose a 
time-dependent denoiser of the forecasted series using prospective sources of context data collected along the horizon of prediction. 

In addition to these contributions, a new class of distance-based forecasters is also introduced, combining the advantages of lazy learning with the advances on the calculus of bary-centers from time series data. We show they are competitive with context-unaware LSTMs, and use them to further confront the performance and behavior of the proposed context-aware LSTMs. 

Results gathered over the Lisbon's public bike sharing system (GIRA) confirm the role of the above contributions, even though not all improvements yield statistically significant differences. 

To our knowledge, this work provides the first attempt to systematically combine diverse sources of context, as well as seize the benefits that can be propelled by the use of both historical and prospective context data. The proposed principles are simplistic, yet effective, thus being now accessible for researchers worldwide as viable techniques to incorporate a wide diversity of available sources of context data in neural-based predictive models. The introduce principles also form a sound and comprehensive basis for the assessment of future context-aware predictors.

\begin{acks}
\small The authors thank \textit{Câmara Municipal de Lisboa} for data provision, support and valuable feedback, particularly \textit{Gabinete de Mobilidade} and \textit{Centro de Operações Integrado}. This work is further supported by national funds through \textit{Fundação para a Ciência e Tecnologia} under project ILU (DSAIPA/DS/0111/2018) and INESC-ID pluriannual (UIDB/50021/2020).
\end{acks}

\bibliographystyle{ACM-Reference-Format}
\bibliography{references}

\end{document}